\documentclass[letterpaper, 10 pt, conference]{./support/ieeeconf}
\usepackage{times}
\usepackage[pdftex]{graphicx}
\usepackage{amsmath,amssymb,amsopn,amstext,amsfonts}
\usepackage{cancel}
\usepackage[space]{cite}
\usepackage{pdfsync}
\usepackage{balance}
\usepackage{color}
\usepackage{mathtools}
\usepackage{algpseudocode}
\usepackage{algorithm} 
\usepackage{bm}

\usepackage{diagbox}
\usepackage{float}
\usepackage{epstopdf}
\usepackage{pifont}
\usepackage{fixltx2e}
\usepackage{amsmath}
\usepackage{multirow}
\usepackage{url}
\usepackage{verbatim}
\usepackage{siunitx} 
\usepackage{epstopdf}

\usepackage[linkcolor=black,citecolor=black,urlcolor=black,colorlinks=true]{hyperref}
\usepackage{booktabs}
\usepackage{graphicx}
\usepackage{makecell}
\usepackage{enumerate}

\makeatletter
\let\NAT@parse\undefined
\makeatother
\usepackage{subfigure}
\usepackage{caption}

\algnewcommand\algorithmicforeach{\textbf{for each}}
\algnewcommand{\algorithmicgoto}{\textbf{go to}}%
\algnewcommand{\Goto}[1]{\algorithmicgoto~\ref{#1}}%
\algdef{S}[FOR]{ForEach}[1]{\algorithmicforeach\ #1\ \algorithmicdo}

\graphicspath{{../figures/}}
\DeclareGraphicsExtensions{.png,.jpg,.eps,.pdf}
\IEEEoverridecommandlockouts
\title{\LARGE \bf The Visual-Inertial-Dynamical Multirotor Dataset}

\author
{
    Kunyi Zhang$^{1,2,*}$, Tiankai Yang$^{1,2,*}$, Ziming Ding$^{1,2}$, Sheng Yang$^{3}$,\\ Teng Ma$^{3}$, Mingyang Li$^{3}$, Chao Xu$^{1,2}$ and Fei Gao$^{1,2}$
	\thanks{This work was supported by the National Key Research and Development Program of China (Grant NO. 2020AAA0108104), Alibaba Innovative Research (AIR) Program, and National natural Science Foundation of China under Grant 62003299.}
	\thanks{\textsuperscript{1}State Key Laboratory of Industrial Control Technology, Institute of Cyber-Systems and Control, Zhejiang University, Hangzhou 310027, China.}
	\thanks{\textsuperscript{2}Huzhou Institute, Zhejiang University, Huzhou 313000, China.}
	\thanks{\textsuperscript{3}Alibaba DAMO Academy Autonomous Driving Lab, Hangzhou 311121, China.}
	\thanks{E-mail: {\{kunyizhang, fgaoaa\}@zju.edu.cn}}
    \thanks{* Equal contributors.}    
}

\begin{document}

\maketitle
\thispagestyle{empty}
\pagestyle{empty}

\begin{abstract}    
	Recently, the community has witnessed numerous datasets built for developing and testing state estimators. 
    However, for some applications such as aerial transportation or search-and-rescue, 
    the contact force or other disturbance must be perceived for robust planning and control, 
    which is beyond the capacity of these datasets. 
    This paper introduces a Visual-Inertial-Dynamical (VID) dataset, 
    not only focusing on traditional six degrees of freedom (6-DOF) pose estimation 
    but also providing dynamical characteristics of the flight platform for external force perception or dynamics-aided estimation. 
    The VID dataset contains hardware synchronized imagery and inertial measurements, 
    with accurate ground truth trajectories for evaluating common visual-inertial estimators. 
    Moreover, the proposed dataset highlights rotor speed and motor current measurements, 
    control inputs, and ground truth 6-axis force data to evaluate external force estimation. 
    To the best of our knowledge, the proposed VID dataset is the first public dataset 
    containing visual-inertial and complete dynamical information in the real world for pose and external force evaluation. 
    The dataset\footnote{\url{https://github.com/ZJU-FAST-Lab/VID-Dataset}\label{web_dataset}}
    and related files\footnote{\url{https://github.com/ZJU-FAST-Lab/VID-Flight-Platform}\label{web_platform}}
    are open-sourced.
\end{abstract}

\IEEEpeerreviewmaketitle

\section{Introduction}
    
As the demand of GPS-denied navigation increases, 
state estimators play an indispensable role in the real-time estimation of accurate robot poses.
However, in many applications, there is an pressing requirement for the estimation of contact forces or other disturbances.
For instance, in aerial transportation and delivery, drones are required to operate under a heavy payload, thus forming a multi-rigid body~\cite{Clark2020realtime}, whose mathematical model needs to be recognized online. 
In aerial manipulation, a drone equipped with a manipulator~\cite{Suseong2018robust,Dongjae2019model} also requires precise force feedback to stabilize itself.
Moreover, in autonomous flight with severe winds~\cite{seo2019robust,ji2020cmpcc,wu2021external}, the absence of an accurate external disturbances estimation significantly harms the effectiveness of planning and control. 
In response to the above, we propose a dataset containing multirotor dynamics as well as visual and inertial data.

In the decade, there are several datasets~\cite{burri2016euroc,majdik2017zurich,sun2018robust,delmerico2019we,antonini2020blackbird} 
for state estimation of aerial robots have been proposed.
These datasets typically provide exteroceptive sensing, 
including imagery and range measurements, and proprioceptive data such as inertial measurements, 
for developing onboard pose estimation algorithms. 
The EUROC dataset~\cite{burri2016euroc} pioneers the visual-inertial benchmarking for drones, 
where a collection of sequences with accurate ground truth is presented. 
Although the EUROC dataset significantly pushes the boundary of visual SLAM, 
it only contains motions with moderate speed, thus not applicable to aggressive flights.
The Zurich Urban MAV~\cite{majdik2017zurich} dataset contains long-distance data 
recorded from a tethered multirotor flying in an urban environment without highly accurate ground truth.
The Upenn Fast Flight~\cite{sun2018robust} provides challenging sequences 
of fast speed and changing illumination but lacks ground truth as well. 
Recently, UZH proposes a first-person-view (FPV) dataset~\cite{delmerico2019we} 
to challenge the capability of existing state estimation algorithms in extreme conditions. 
This dataset consists of visual, event, and inertial streams collected in aggressive flights. 
Besides, the Blackbird Dataset~\cite{antonini2020blackbird} also targets high-speed applications. 
It provides fruitful scenes with photorealistic image rendering and also includes motor speed sensors. 

\begin{figure}[t]
    \vspace{0cm}
    \begin{center}
        \includegraphics[angle=0,width=0.45\textwidth]{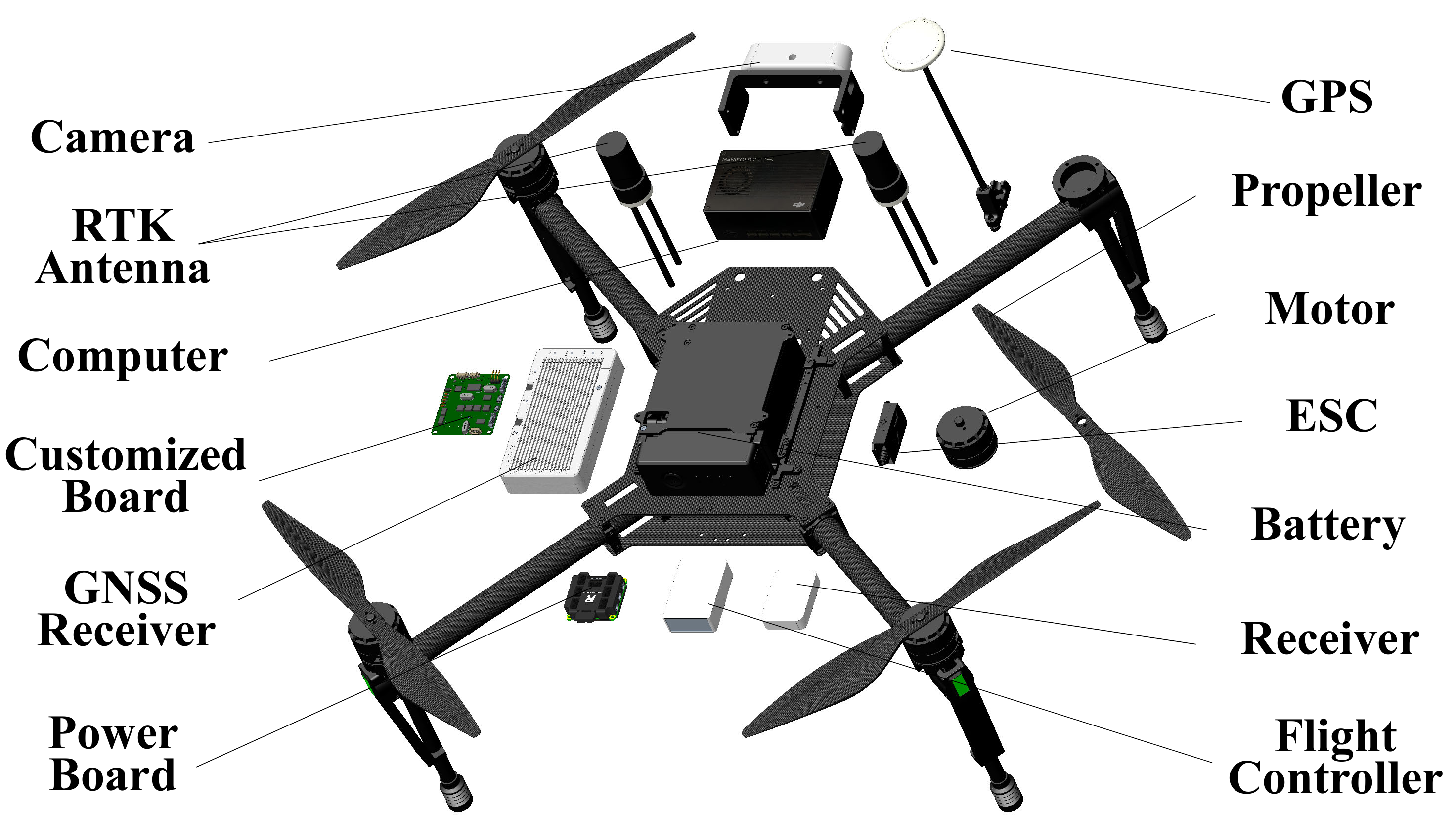}
        \caption{Multirotor platform built for the dataset}
        \label{assembly}
    \end{center}
    \vspace{-1.7cm}
\end{figure}

\begin{table*}[h]
	\centering
	\begin{tabular}{ccccccc}
		
		\toprule
		& \makecell[c]{EuRoC\\MAV           \\ \cite{burri2016euroc}}
		& \makecell[c]{Zurich\\Urban MAV    \\ \cite{majdik2017zurich}}
		& \makecell[c]{UPenn\\Fast Flight 	\\ \cite{sun2018robust}} 
		& \makecell[c]{UZH FPV\\dataset 	\\ \cite{delmerico2019we}}
		& \makecell[c]{Blackbird\\dataset 	\\ \cite{antonini2020blackbird}}
		& \textbf{Ours}	                    \\
		\toprule
		Camera$^{a}$   				& 20$Hz$  	& 20$Hz$    & 40$Hz$    & 30/50$Hz$		& 120/60$Hz^{e}$  	& 60$Hz^{Synced}$  \\
		IMU  						& 200$Hz$   & 10$Hz$   	& 200$Hz$   & 500/1000$Hz$  & 100$Hz$     		& 400$Hz^{Synced}$      \\
		Segmentation  				& n/a   	& n/a       & n/a       & n/a           & 60$Hz$      		& n/a                   \\
		Ground truth  				& 100$Hz$	& n/a       & n/a       & 20$Hz$		& 360$Hz$			& 100/5$Hz$               \\
		RTK							& n/a		& n/a       & n/a       & n/a 		    & n/a				& 5$Hz$                 \\
		Control inputs$^{b}$		& n/a		& n/a       & n/a       & n/a 		    & ~190$Hz$			& 100$Hz^{Synced}$      \\
		Tachometer$^{c}$			& n/a   	& n/a       & n/a       & n/a           & ~190$Hz$    		& 1000$Hz^{Synced}$	    \\
		Ammeter$^{c}$				& n/a   	& n/a       & n/a       & n/a           & n/a       		& 1000$Hz^{Synced}$     \\
		6-axis force sensor$^{d}$	& n/a       & n/a       & n/a       & n/a			& n/a       		& 100$Hz$               \\ 
		\bottomrule
	\end{tabular}
	
	\caption{Multirotor datasets comparison. 
		${^\emph{a}}$ Frequencies are expressed as (general image)/(range image). 
		${^\emph{b}}$ Control inputs mean the veritable inputs to motors, including target rotor speeds and motor currents.
		${^\emph{c}}$ Both are measurements of rotor speed and motor current. 
		${^\emph{d}}$ Sensor data from a 6-axis force sensor. 
		${^\emph{e}}$ Visual environments are synthesized in photorealistic simulation.
		${^\emph{Synced}}$ Sensor data are time-synchronized in the VID dataset.    
	}
	\label{dataset_compare}
	\vspace{-1.0cm}
\end{table*}

Thanks to the above-mentioned public datasets, state estimation algorithms have been significantly improved in the last decade. 
The early visual odometry systems, such as SVO~\cite{forster2014svo}, ORB-SLAM~\cite{mur2015orb} and DSO~\cite{engel2017direct} 
estimate 6-DoF poses with merely images. 
Modern IMU-aided visual odometry systems, 
including MSCKF~\cite{li2013high}, Vins-Mono~\cite{qin2018vins}, OpenVINS~\cite{geneva2020openvins} and ORB-SLAM3~\cite{campos2021orb},
opt to utilize accelerometer and gyroscope as kinematic inputs to get more accurate state estimation.
Besides, model-based visual-inertial odometry VIMO~\cite{nisar2019vimo} and visual-inertial-dynamics state estimation VID-Fusion~\cite{ding2021vid}, which simultaneously estimate the pose and the external force, are a trending research topic emergent in the recent SLAM community. 
Since a multirotor's dynamical characteristics can be determined entirely from the rotor model~\cite{mahony2012multirotor}, dynamics is considered as a new information source that can further help state estimation, especially for highly aggressive motions.
These estimators also obtain the external force, torque, or disturbance applied to the vehicle in real-time as a byproduct.

Nevertheless, few datasets focus on the dynamical characteristics of multirotor platforms to support the aforementioned researches. 
The Blackbird dataset~\cite{antonini2020blackbird} takes one step towards this by providing a model of multirotor dynamics, which is simply described as the sum of all motor speeds and their corresponding thrusts.
There are four additional motor speed sensors (as tachometers) embedded in the multirotor with customized optical motor encoders, and the thrust coefficient measured from experiments conducted in a wind tunnel.
However, there are some drawbacks existing in the Blackbird dataset. 
Firstly, only one propeller thrust coefficient is provided to represent all.
Actually for a multirotor, the thrust coefficients usually vary (as in TABLE~\ref{dynamic_factor}).
Secondly, the Blackbird dataset is insufficient for interactive perception of the environment due to the absence of ground truth of external force.

\begin{figure}[t]
	\vspace{0.5cm}
	\begin{center}
		\includegraphics[angle=0,width=0.5\textwidth]{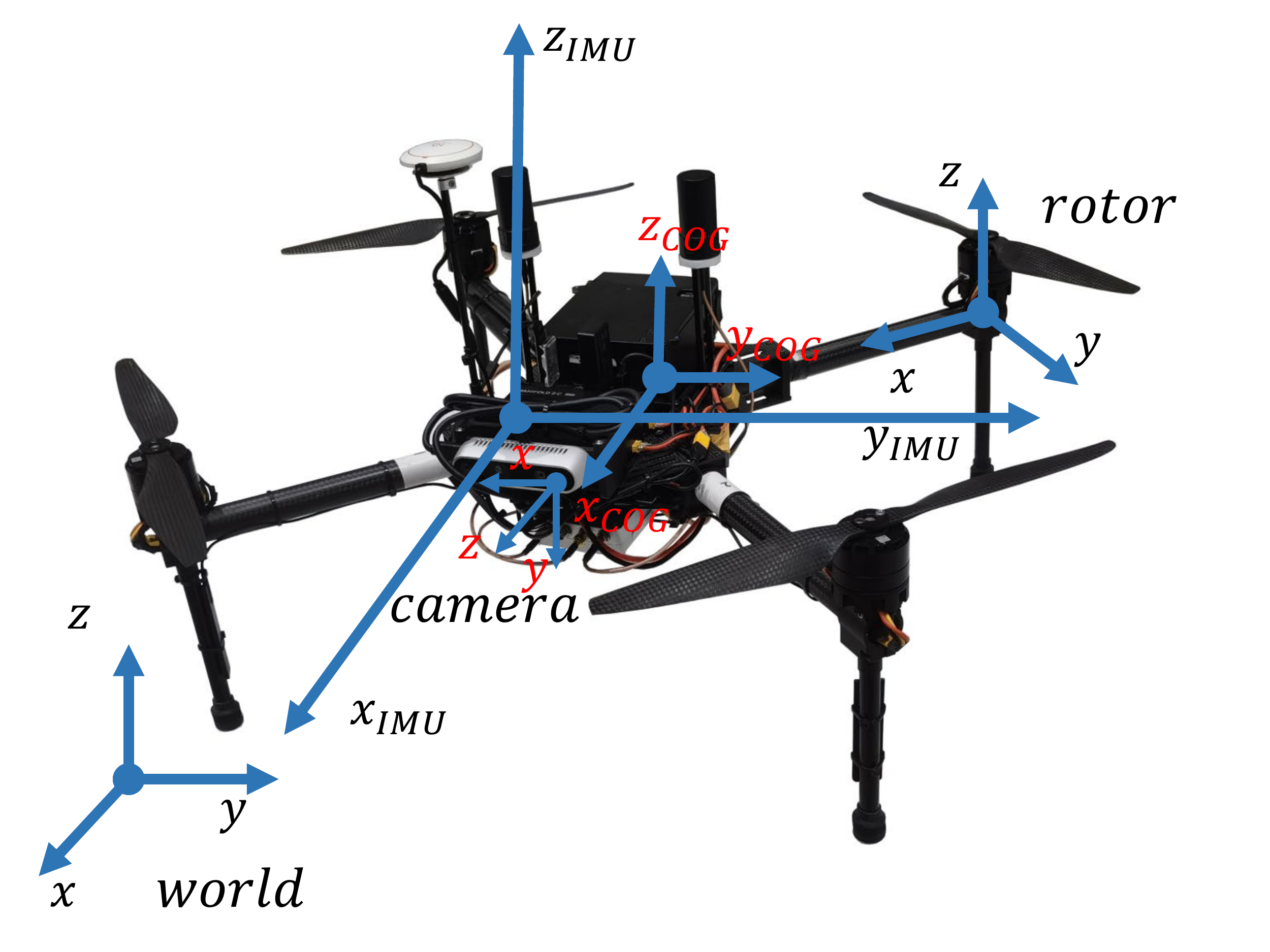}
		\caption{Flight platform with coordinates.}
		\label{coordinates} 
	\end{center}
	\vspace{-1.2cm}
\end{figure}

To bridge this gap and lay the foundation for developing a dynamics-aided state estimator for multirotors, we herein present the VID dataset with complete dynamical characteristics.
The distinctive features of the VID dataset compared with others are demonstrated in TABLE~\ref{dataset_compare}.
Besides, there are several unique contributions of this work:
\begin{enumerate}
	\item An available and completely public dataset for pose and external force estimation, which includes visual, inertial and dynamical data as well as the ground truth of external forces and poses.
	\item An open-sourced flight platform equipped with versatile sensors for more advanced applications, along with fully identified dynamical and inertial parameters.
	\item A general and exhaustive hardware synchronization solution for time alignment of images, inertial measurements, control inputs, rotor speeds, and motor currents, which leads to better sensor fusion.
\end{enumerate}

\section{Datasets}
\label{sec:dataset}
\subsection{Flight platform}
\begin{figure*}[t]
	\vspace{0.0cm}
	\begin{center}
		\includegraphics[angle=0,width=0.9\textwidth]{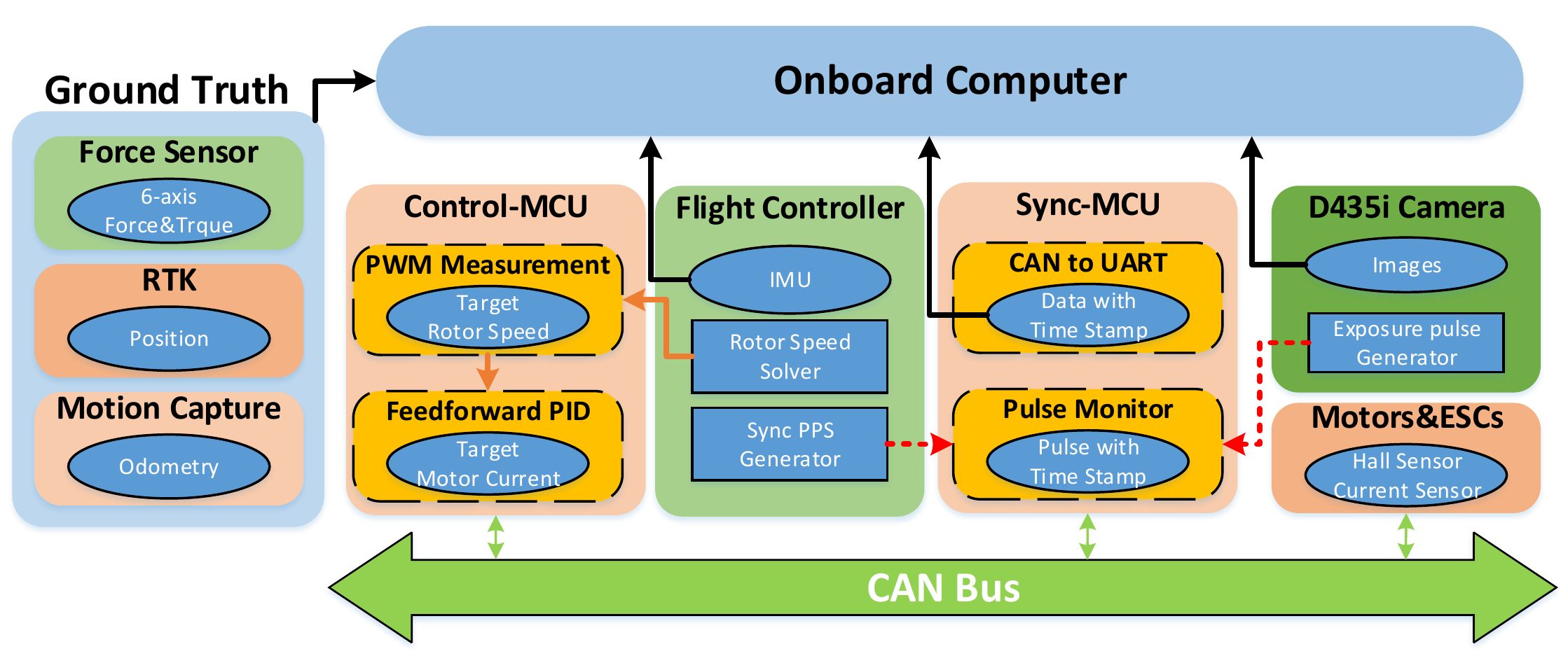}
		\caption{Hardware system framework (Red lines: sync pluses)}
		\label{hardware_diagram} 
	\end{center}
	\vspace{-0.8cm}
\end{figure*}

The platform (as in Fig.~\ref{assembly}) is modified from a 650mm wheelbase multirotor\footnote{\url{https://www.dji.com/cn/matrice100}}.
To get high-precision measurements of rotor speed and motor current, 
a motor with a pair of orthogonal Hall position sensors and a current sampling unit\footnote{\url{https://www.robomaster.com/zh-CN/products/components/general/M3508}} is selected as the propulsion unit.
There are
a depth camera\footnote{\url{https://www.intelrealsense.com/depth-camera-d435i}}, 
a IMU integrated with the flight controller\footnote{\url{https://www.dji.com/cn/n3}},
a customized circuit board\textsuperscript{\ref{web_platform}} installed with 2 microcontrollers, separately for motor control (Ctrl-MCU) and hardware synchronization (Sync-MCU), 
and a computer\footnote{\url{https://www.dji.com/cn/manifold-2}} for data collection
equipped on the multirotor.
Here, we list common parameters of the multirotor and onboard sensors in TABLE~\ref{parameters} 
and show the relevant coordinates in Fig.~\ref{coordinates}.
The hardware system framework is shown in Fig.~\ref{hardware_diagram}.

\subsection{Data sources}
\subsubsection{Rotor speed and motor current}
The motor feedbacks rotor speed and current messages with the Controller Area Network (CAN) at 1000$Hz$.
The Sync-MCU monitors the above messages and sends them to the onboard computer by a Universal Asynchronous Receiver Transmitter (UART).
The units of rotor speed and current are revolutions-per-minute ($rpm$) and $\frac{{20}}{{{2^{16}}}}A$.

\subsubsection{Control input}
The flight controller continuously generates control commands and sends them out by Pulse Width Modulation (PWM) waves.
The Ctrl-MCU measures the waves to get 100$Hz$ target rotor speeds
and uses a PID regulator with feedforward compensation to calculate target motor currents at 100$Hz$.
Then, the Ctrl-MCU sends the control inputs (target rotor speeds and motor currents) to the CAN bus, and the Sync-MCU converts the above CAN messages to UART packs and sends them to the computer.

\subsubsection{IMU}
The IMU measures acceleration, angular velocity and magnetometer data at 400$Hz$.

\subsubsection{Camera}
The camera offers stereo imagery and range measurements at 60$Hz$.

\subsection{Hardware time synchronization}

To improve the quality of sensor fusion, the above onboard sensors are hardware synchronized by electronic pulses.
Fig.~\ref{hardware_diagram} illustrates the relevant devices, pulses and dataflow.
When the time synchronization system starts to work,
the flight controller generates a pulse-per-sec (PPS) and marks the IMU data corresponding to the pulse (one pulse every 400 IMU data).
The camera generates a 60$Hz$ pulse, which corresponds to the time when the camera starts to expose.
The Sync-MCU monitors both above pulses and stamps them with its internal time.
Besides, the Sync-MCU stamps the rotor speeds and motor currents from the CAN bus when receiving them.
After the pulse messages are checked and processed, the data from different reference systems are aligned to the system.

\subsection{Ground truth}
\subsubsection{Indoor pose}
In an ${18m\times9m\times3.5m}$ indoor environment,
we use a motion capture system\footnote{\url{https://www.vicon.com/hardware/cameras/vantage}}
composed of 15 cameras to provide the multirotor's 6-DOF pose at $100Hz$.
\subsubsection{Outdoor pose}
\begin{figure}[t]
	\vspace{0.25cm}
	\begin{center}
		\includegraphics[angle=0,width=0.45\textwidth,height=0.45\textwidth]{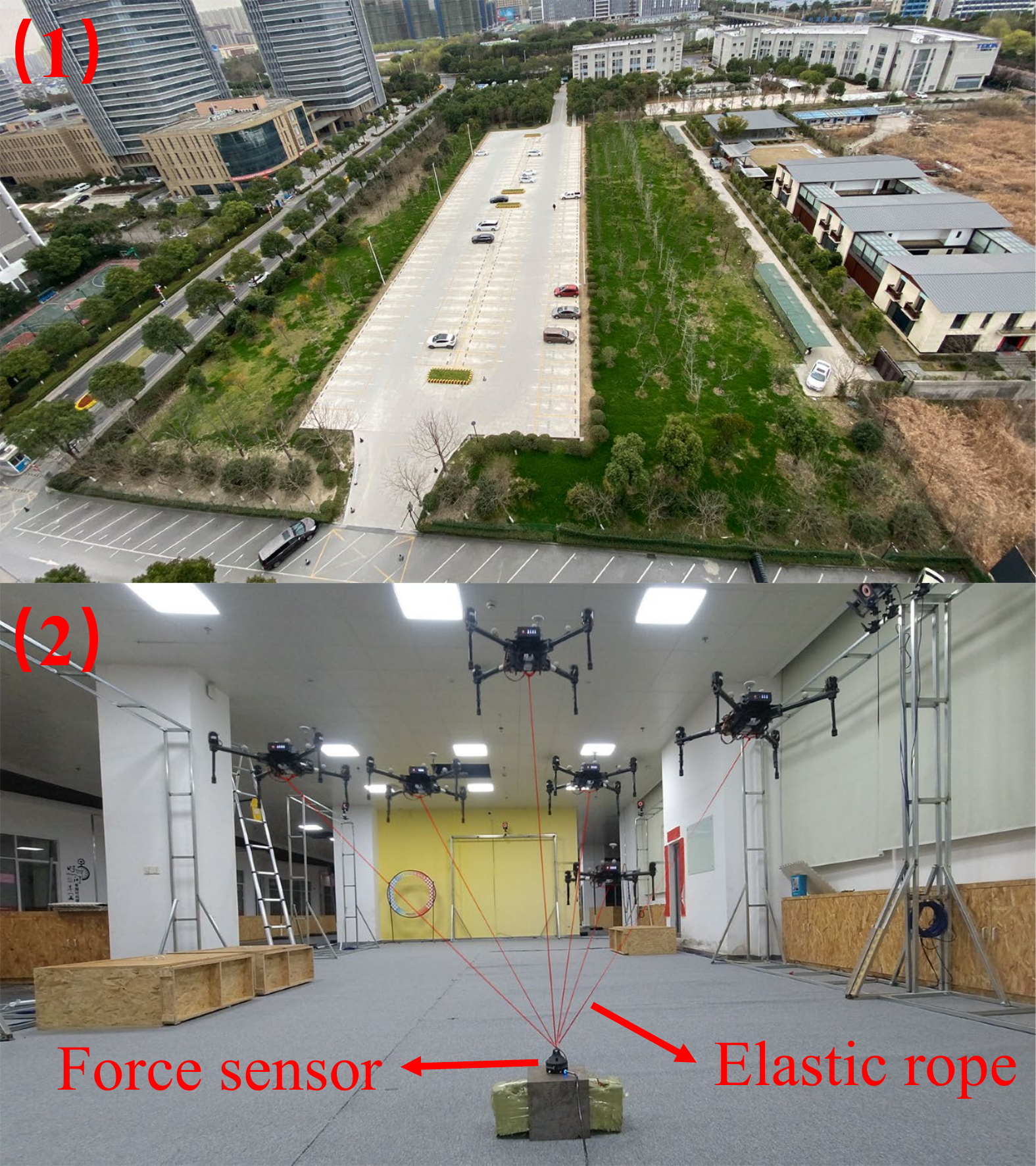}
		\caption{Experiment place.
			(1) Outdoor scene in a parking lot.
			(2) Indoor scene when recording data.
		}
		\label{experiment_place} 
	\end{center}
	\vspace{-2.0cm}
\end{figure}

In a ${100m\times35m}$ outdoor parking environment (Fig.~\ref{experiment_place}),
the multirotor is equipped with a Global-Navigation-Satellite-System (GNSS) receiver\footnote{\url{https://www.zhdgps.com}},
which subscribes to a Real-Time Kinematic (RTK) location service\footnote{\url{https://mall.qxwz.com/market/services/FindCM}}
and receives the multirotor's pose at $5Hz$. 

\subsubsection{External force}
As shown in Fig.~\ref{experiment_place}, 
we fix a 6-axis force sensor\footnote{\url{https://robotiq.com/products/ft-300-force-torque-sensor}} on the ground.
Indoors, we use an elastic rope with one end fixed on the force sensor and the other end pulling the multirotor to obtain the ground truth of the external forces acting on the multirotor during flight.


    


\begin{figure}[t]
   	\vspace{0.0cm}
   	\begin{center}		\includegraphics[angle=0,width=0.45\textwidth]{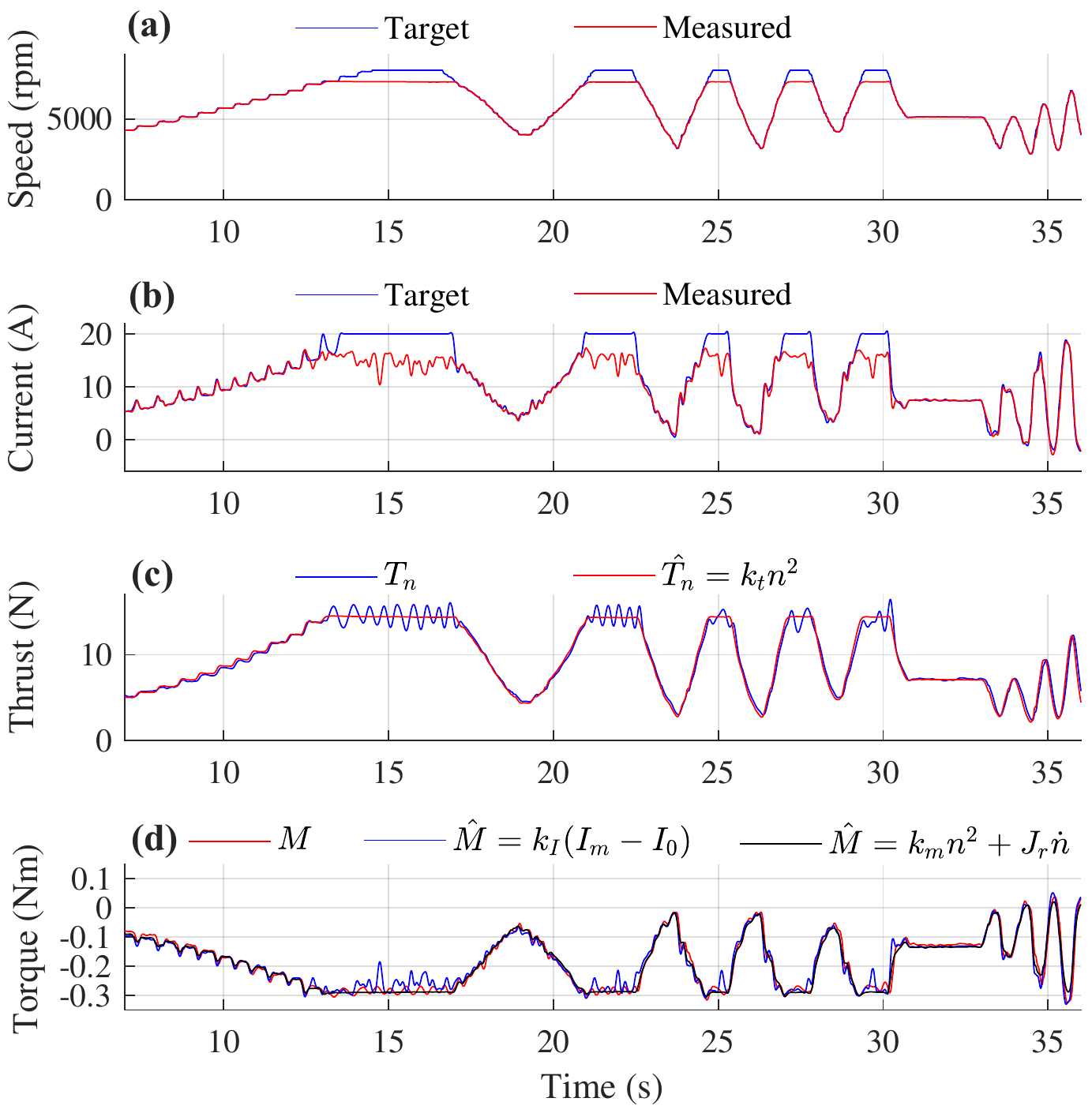}
   		\caption{Propulsion system unit test (Seq.~19).
   			(1) Target rotor speed and measured rotor speed;
   			(2) Target motor current and measure motor current;
   			(3) The thrust measured by the force sensor and the thrust fitted from rotor speed;
   			(4) The torque measured by the force sensor and the torque fitted by rotor speed and motor current.
   		}
   		\label{propeller_dynamics_test_1} 
   	\end{center}
   	\vspace{-0.8cm}
\end{figure}

\begin{table*}[t]
	\centering
	\begin{tabular}{llllll}
		\toprule
		Prop. 					& Value 											& Description &
		Prop. 					& Value 											& Description \\
		\toprule
		$l$					  	& 650 $mm$                                          & Distance wheelbase &
		propeller				& 13 $\times$4.4 $inch$                             & Propeller model\\
		$\eta_{n}$				& 1 $rpm$                                           & Resolution of tachometer$^{a}$ &
		$\sigma_{n}$			& 1.459 $rpm/\sqrt{Hz}$                             & Noise density of tachometer$^{a}$ \\
		$\eta_{I}$				& 1.2207${e^{ - 3}}A$                               & Resolution of ammeter$^{a}$ &
		$\sigma_{I}$			& 4.950${e^{ - 2}}A/\sqrt {Hz} $                    & Noise density of ammeter$^{a}$ \\
		$\sigma_{g}$			& 0.001 $rad/s/\sqrt{Hz}$    						& Noise density of gyr.$^{b}$ &
		$\sigma_{b_g}$			& 3.500e-5 $rad/s^2/\sqrt{Hz}$   					& Gyr. random walk bias$^{b}$ \\
		$\sigma_{a}$			& 0.100 $m/s^2/\sqrt{Hz}$     						& Noise density of acc.$^{b}$ & 
		$\sigma_{a_g}$			& 4.000e-3 $m/s^3/\sqrt{Hz}$						& Acc. random walk bias$^{b}$ \\
		$\eta_{f}$  			& [0.1, 0.1, 0.1] $N$    							& Noise of force sensor$^{c}$ &
		$\eta_{m}$ 	 			& [5e-3, 5e-3, 3e-3] $N\cdot m$ 					& Noise of torque sensor$^{c}$ \\
		$T{_I^C}$				& [1.8875e-1, 1.75e-3, 5.55e-2] $m$     			& Trs. IMU $\rightarrow$ camera$^{d,e}$ &
		$T{_I^G}$				& [0.0, 0.0, 9.25e-3] $m$							& Trs. IMU $\rightarrow$ geometric$^{d,f}$ \\
		Image size				& 640px$\times$480px								& Image width and height &
		baseline				& 50$mm$											& Distance between stereo pair \\
		\bottomrule
	\end{tabular}
	\caption{Multirotor characteristics. 
		${^\emph{a}}$ Characteristics of the rotor tachometers and the motor ammeters are statistically calibrated. 
		${^\emph{b}}$ IMU calibration result by Allan covariance. 
		${^\emph{c}}$ Noise of force sensor parameters are tested by static experiments. 
		${^\emph{d}}$ ${T_A^B}$ represents translation from A frame to B frame in A frame.
		${^\emph{e}}$ Camera extrinsic translation parameter from CAD model. 
		${^\emph{f}}$ We define the center of the multirotor as the geometric center with the same direction as the IMU frame, which is obtained from the CAD model. }
	\label{parameters}
	\vspace{-0.5cm}
\end{table*}

\section{Data validation}
\label{sec:data validation}

\subsection{Dynamical system modelling}
\label{Dynamics model}	
In order to establish the complete dynamics of the multirotor, 
a basic aerodynamic modelling of a single propeller propulsion system is required.
Here, the Rankine-Froude momentum theorem~\cite{leishman2016principles} 
models the steady-state thrust $T$ (unit: $N$) and counter torque $M$ (unit: $N\cdot m$) generated by a hovering rotor, as shown below:
\begin{equation}
\begin{aligned}
\label{equation_thrust_torque}
	T &:= C_{T}\rho A R^2 \omega^2,\\
	M &:= C_{M}\rho A R^3 \omega^2,
\end{aligned}
\end{equation}
where, for any rotor, ${C_{T}}$ is the thrust coefficient,
${C_{M}}$ is the torque coefficient.
${\rho}$ is the density of air, 
${A}$ is the rotor disk area,
${R}$ is the radius of rotor and ${\omega}$ is the angular velocity.
${C_{T}}$ and ${C_{M}}$ depend on rotor geometry and profile. 
We replace angular velocity ${\omega}$ (unit: ${rad/s}$) with the rotor speed ${n}$ (unit: ${rpm}$), and briefly express Eq.~(\ref{equation_thrust_torque}) as the following lumped parameters:
\begin{equation}
\begin{aligned}
\label{equation_thrust_torque_rpm}
	T &= k{_t}n{^2},\\
	M &= k{_m}n{^2}+J{_r}\dot{n},
\end{aligned}
\end{equation}
where, $k{_t}$, $k{_m}$ and $J{_r}$ mean thrust coefficient, counter torque coefficient and rotor's moment of inertia of z-axis, respectively. It should be noted that, compared with Eq.~(\ref{equation_thrust_torque}), Eq.~(\ref{equation_thrust_torque_rpm}) adds the additional torque generated when the rotor accelerates and decelerates~\cite{Svacha2020IMU}.

In order to identify the above models, we mount a motor fixed with a propeller to the 6-axis force sensor, and obtain the model parameters by analyzing its tachometer and thrust data.
Fig.~\ref{propeller_dynamics_test_1} shows the aerodynamics of a propulsion unit, the complete identification parameters are listed in the TABLE~\ref{dynamic_factor}. 
The thrust coefficients of the four channels are similar, but the torque coefficients differ significantly, especially between the clockwise and counterclockwise propellers.

\begin{table}[t]
	\vspace{0.1cm}
	\centering
	\begin{tabular}{lcccc}
		\toprule
		& \makecell[c]{$k{_t}$ \\ $(N/rpm^2)$}      							& \makecell[c]{$k{_m}$ \\ $(N\cdot m/rpm^2)$}			
		& \makecell[c]{$J{_{r}}$ \\ $(kg\cdot m^2)$}							&\makecell[c]{$k{_I}$ \\ $(N\cdot m/A)$}\\
		\toprule
		$1$  			& 2.6890$e$-7   			& -5.6343$e$-9				& -9.8316$e$-6				& -2.7506$e$-5\\
		$2$ 			& 2.8190$e$-7    			&  4.7180$e$-9				&  8.5648$e$-6   			&  2.2204$e$-5\\
		$3$ 			& 2.7263$e$-7    			& -5.7012$e$-9  			& -9.7166$e$-6				& -2.7694$e$-5\\
		$4$ 			& 2.7741$e$-7    			&  4.8260$e$-9				&  9.8131$e$-6				&  2.1910$e$-5\\
		\bottomrule
	\end{tabular}
	\caption{Propeller dynamics model. The number represents the designed multirotor's rotor or motor channel, counting from the upper right corner counterclockwise. A negative sign means the opposite direction.} 
	\vspace{-1cm}
	\label{dynamic_factor}
\end{table}
	
\begin{figure}[t]
	\vspace{0.1cm}
	\begin{center}		
		\includegraphics[angle=0,width=0.45\textwidth]{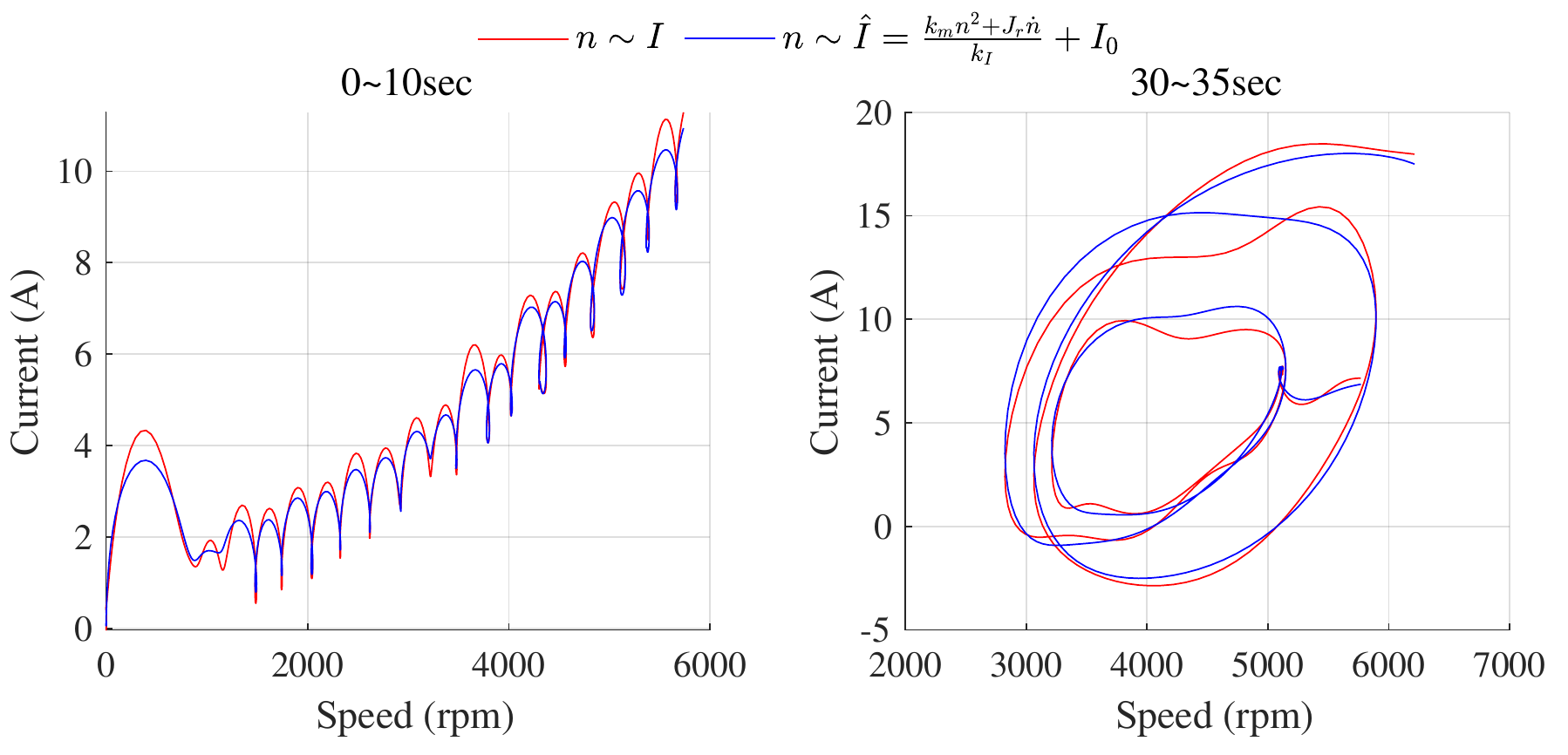}
		\caption{Comparison of the relationship between rotor speed and current with measured value and fitting model (Seq.~19).}
		\label{propeller_dynamics_test_2} 
	\end{center}
	\vspace{-1.7cm}
\end{figure}

Apart from rotor speed ${n}$, motor armature current $I{_m}$ is also provided for redundant input resource. 
The counter torque provided by the rotor is equivalent to the output torque of the motor, and according to electric machinery~\cite{chapman2012electric}, the latter is proportional to the armature current. We can build another torque model as follows:
\begin{equation}
\label{equation_torque_current}
	\begin{aligned}
		M 	&= M{_e} - M{_0} \\
		&= k{_{I}} (I{_m} - I{_0}),
	\end{aligned} 
\end{equation}
where $k{_I}$ is the current torque coefficient. $I{_0}$ and $M{_0} =  k{_I}I{_0}$ represent actual no-load current and no-load torque, respectively. 
In other words, we can only drive the rotor to rotate if armature current $I{_m}$ is greater than $I{_0}$.
The forth chart of Fig.~\ref{propeller_dynamics_test_1} also adds a comparison with Eq.~\ref{equation_torque_current}. 
The corresponding current torque coefficients of the multirotor are also shown in TABLE~\ref{dynamic_factor}.
Besides, the relationship of rotor speed and motor current can be derived from Eq.~\ref{equation_thrust_torque_rpm} and \ref{equation_torque_current} and formulated as follow:
\begin{equation}
	\label{equation_rpm_current}
	\begin{aligned}
		I_m = \frac{k{_{m}} n^2 + J{_r}\dot{n}}{k{_{I}}} + I{_0},
	\end{aligned} 
\end{equation}
which is shown in Fig.~\ref{propeller_dynamics_test_2}.

\subsection{Rotor speed and motor current}
For more rapid motor response, 
we use corresponding motor current calculated by Eq.~\ref{equation_rpm_current} to output to the ESC,
rather than controlling the motor directly by PWM on our platform.
In order to verify the accuracy of the rotor speed and motor current and the validity of dynamical model of the propulsion unit,
we compare target rotor speed and its corresponding motor current with their measured values in the same sequence demonstrated in Fig.~\ref{propeller_dynamics_test_1}.
As shown, the actual (measured) values can closely track the target values by the aforementioned method. In fact, however, neither of them can reach the theoretical maximum due to the insufficient driving forces of the ESCs.
By the way, the measured motor current data is drawn after filtering due to the high noise level.

\subsection{Calibration}

In order to make better use of the sensor data,
it is of great importance to get
the characteristics of the sensors themselves, 
the extrinsic parameters between different sensors 
and the inertial parameters of the flight platform.

The extrinsic parameters between the IMU and the camera are calibrated by Kalibr~\cite{Furgale2013Unified} or measured from CAD drawings.
IMU intrinsic parameters are calibrated by Kalibr-allan\footnote{\url{https://github.com/ethz-asl/kalibr/wiki/IMU-Noise-Model}}.
Besides, we also quantitatively analyze the noise characteristics of the rotor tachometers and current ammeters.
More detailed information can be found on the website\textsuperscript{\ref{web_dataset}}.

\begin{figure}[h]
	\vspace{0.0cm}
	\begin{center}
		\includegraphics[angle=0,width=0.5\textwidth]{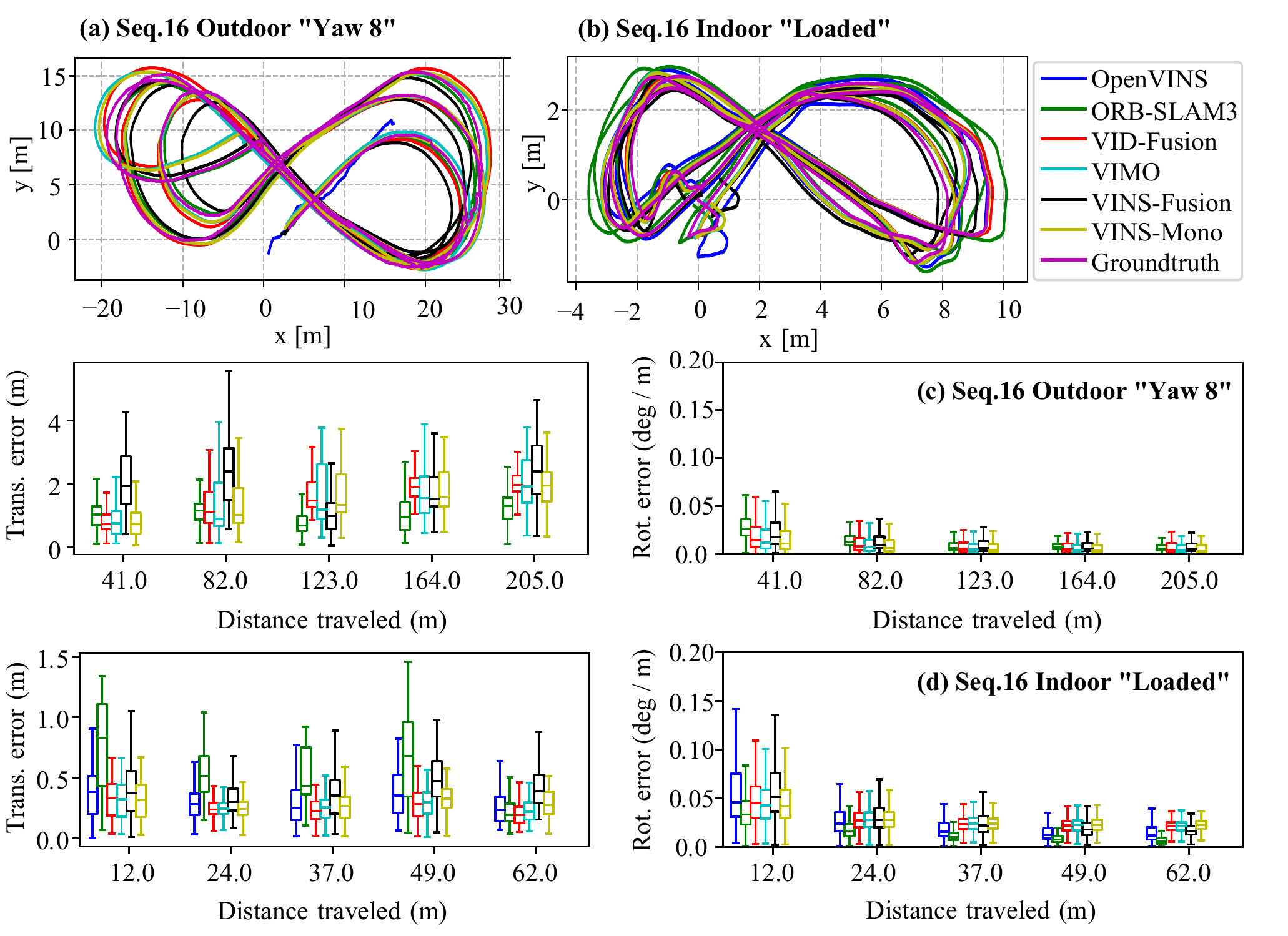}
		\caption{Pose estimation of various algorithms in different sequences.}	
		\label{pose_estimation}
	\end{center}
	\vspace{-2.5cm}
\end{figure}

\section{Experiments}
\label{sec:experiments}

\subsection{Pose estimation}
Robust and accurate pose estimation is beneficial for external force estimation,
so we select two scenarios to show the accuracy of some state-of-the-art algorithms before external force estimation. 
OpenVINS, ORB-SLAM3 and VINS-Fusion work with binocular images and inertial measurements, Vins-Mono takes monocular images and inertial measurements, VIMO and VID-Fusion add some dynamical model and rotor speed compared to Vins-Mono. 
The last two regard the multirotor as a mass point model, so we input the algebraic sum of the 4 propeller thrusts (as described in \ref{Dynamics model}) as the driving force to the algorithms.
It is worth mentioning that, all algorithms take the same camera intrinsic, extrinsic parameters, imu parameters, and dynamics parameters with loop-closure disabled.

As shown in Fig.~\ref{pose_estimation}, OpenVINS, Vins-Mono, VIMO and VID-Fusion outperform over the other two algorithms, 
yet VIMO and VID-Fusion with extra dynamical information do not show a significant improvement over Vins-Mono in indoor sequences.
ORB-SLAM3 works best outdoors, while OpenVINS hangs due to the presence of a moving vehicle in the initial field of view.
Evaluation results are provided by~\cite{Zhang18iros}.
The VID dataset is competent for the evaluation of vision-based pose estimation algorithms,
and we choose Vins-Mono, which works relatively well indoors, to do a comparative test in external force estimation.

\subsection{External force estimation}
\begin{figure}[t]
	\vspace{0cm}
	\begin{center}
		\includegraphics[angle=0,width=0.48\textwidth]{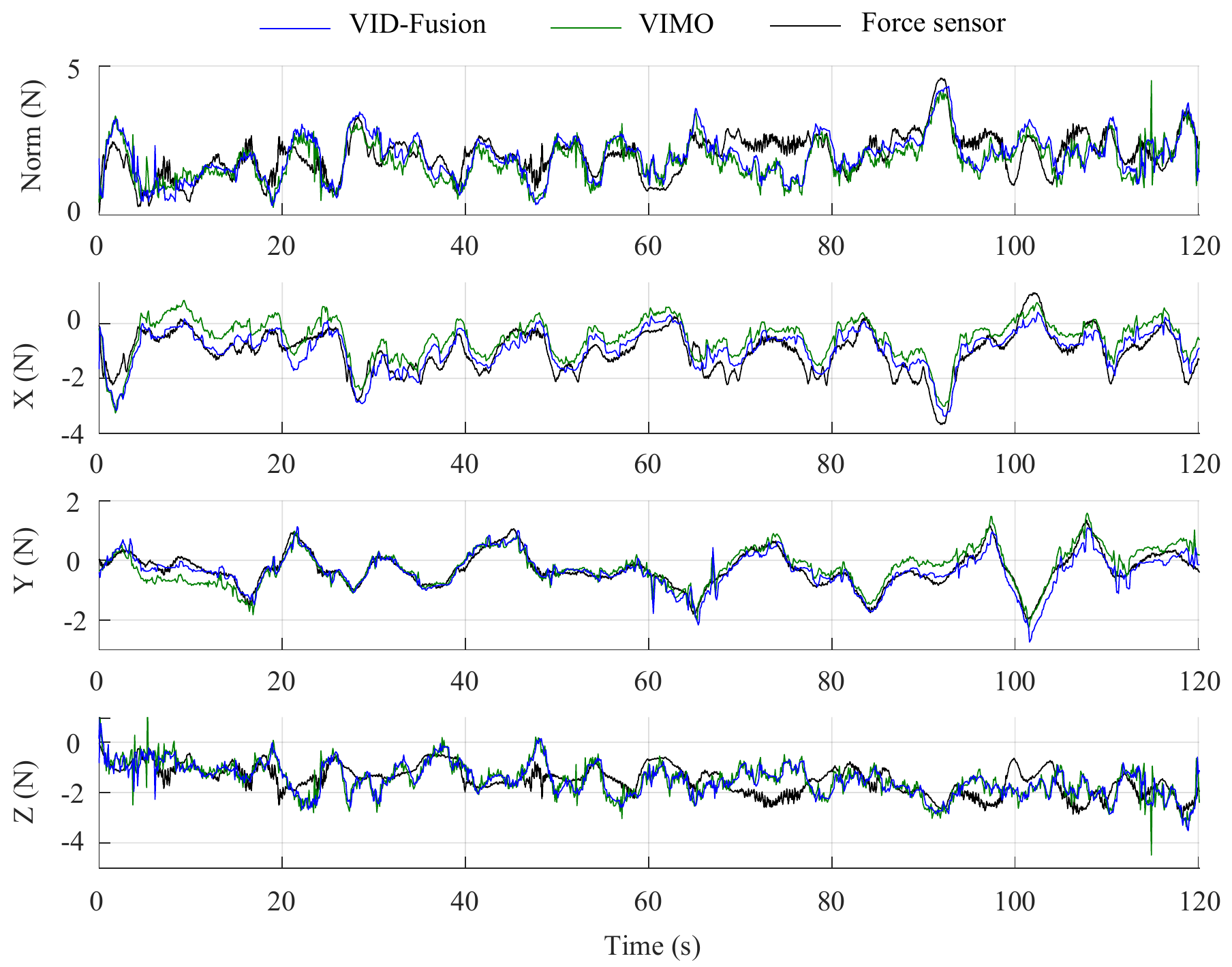}
		\caption{External force estimation with a pulling rope.}
		\label{external_force_estimation_sensor}
	\end{center}
	\vspace{-0.5cm}
\end{figure}

\begin{figure}[h]
	\vspace{0.0cm}
	\begin{center}
		\includegraphics[angle=0,width=0.48\textwidth]{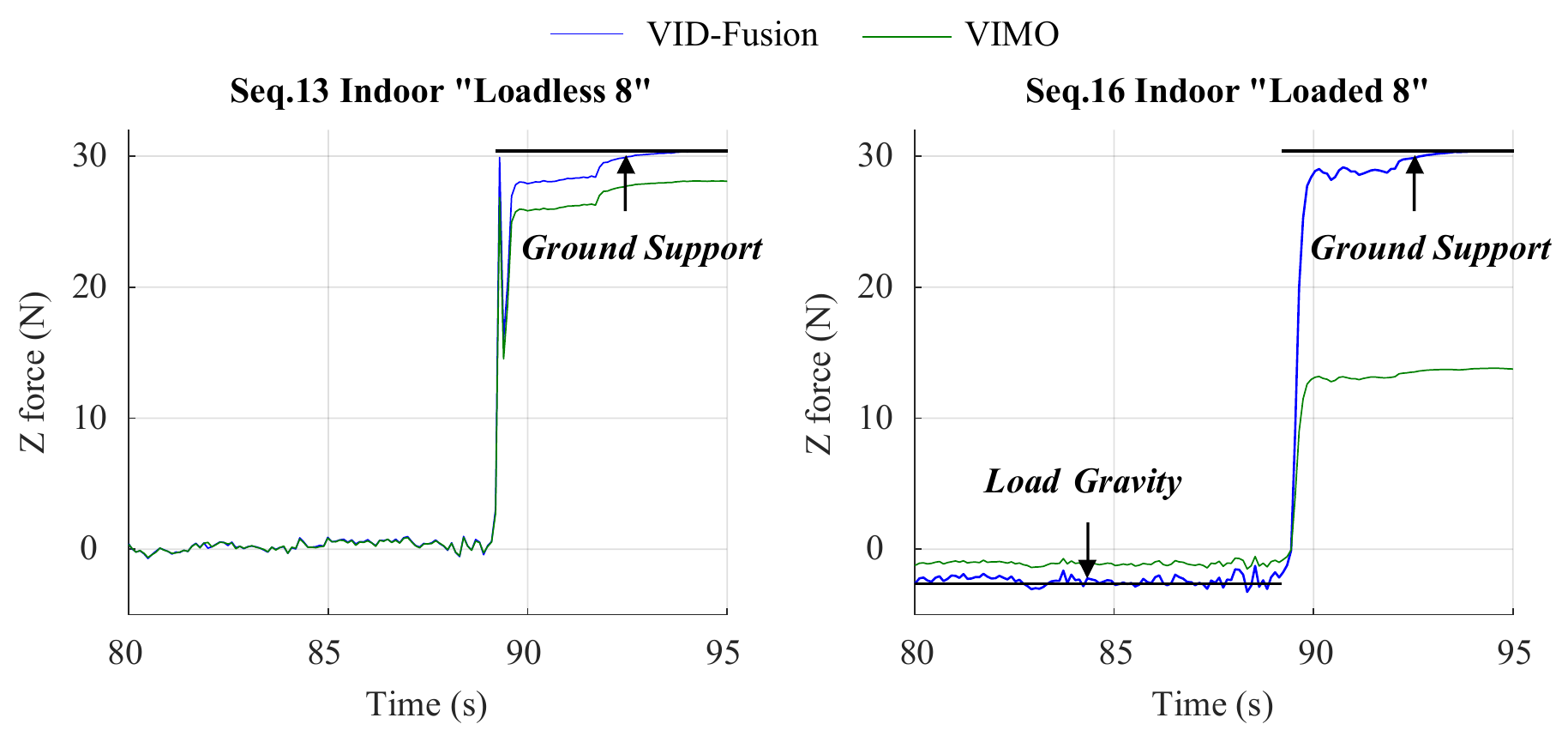}
		\caption{External force estimation with loads.}
		\label{external_force_estimation_load}
	\end{center}
	\vspace{-0.8cm}
\end{figure}

The external force estimation performance of VID-Fusion and VIMO is evaluated in three scenarios 
including 
flights without load (seq.~13), with load (seq.~16) and with a elastic rope (seq.~17).
Fig.~\ref{external_force_estimation_load} illustrates that both algorithms can sense and estimate the weight of the load during flight and the ground support after landing, which corresponds to the multirotor weight withoutload.
But VIMO can not estimate the force correctly after the multirotor is loaded,
mainly because it models the external force as a zero-mean noise.
Fig.~\ref{external_force_estimation_sensor} demonstrates the external force estimation from the rope pulling experiment (seq.~17).
Both algorithms can barely track the change in the contact force (the force transmitted by the rope),
but also have errors in the z-axis.
\begin{figure}[t]
	\vspace{0.0cm}
	\begin{center}
		\includegraphics[angle=0,width=0.48\textwidth]{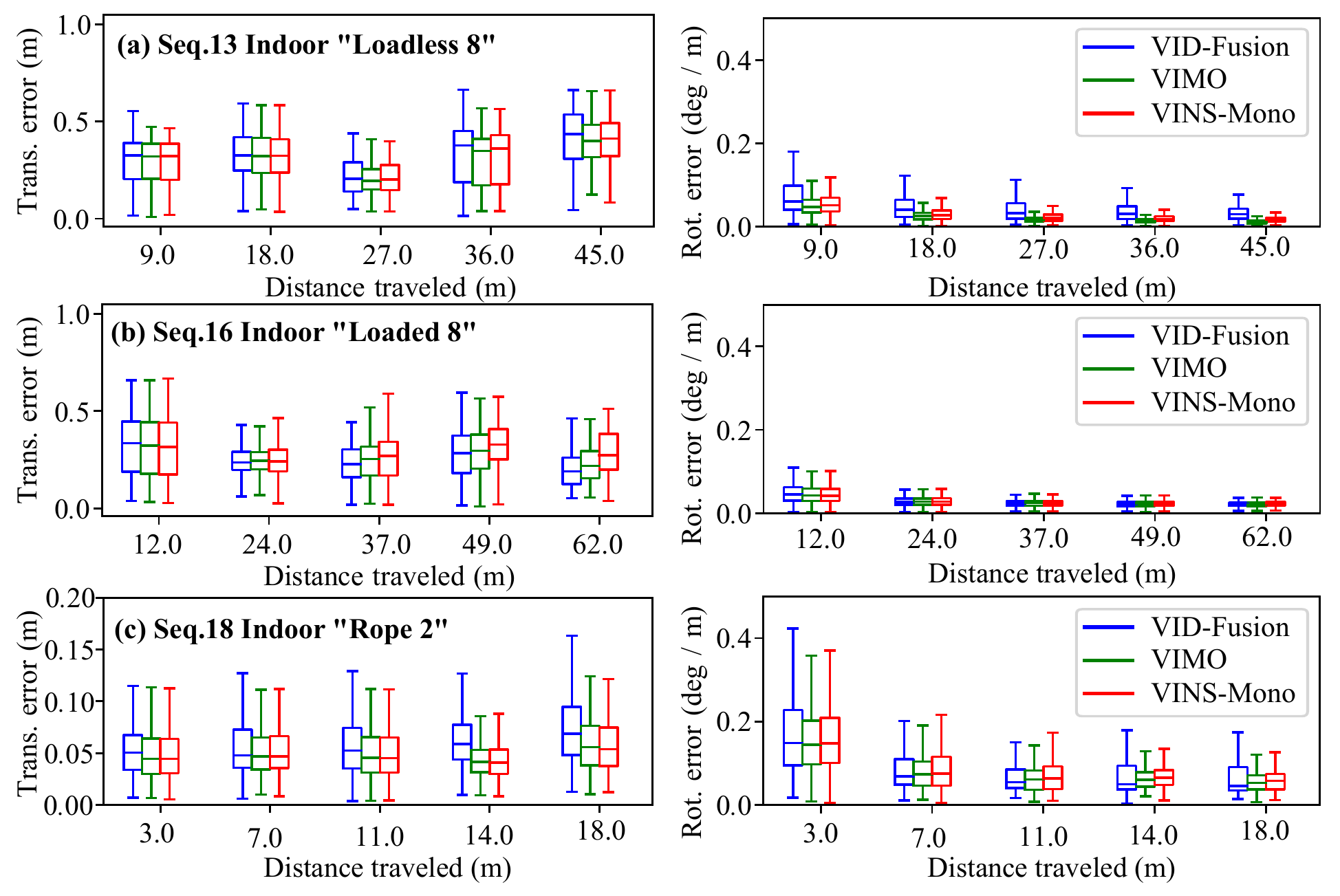}
		\caption{Pose estimation of VID-Fusion, VIMO and Vins-Mono.}
		\label{pose_estimation_3}
	\end{center}
	\vspace{-0.5cm}
\end{figure}
\begin{figure}[t]
	\vspace{0.0cm}
	\begin{center}
		\includegraphics[angle=0,width=0.48\textwidth]{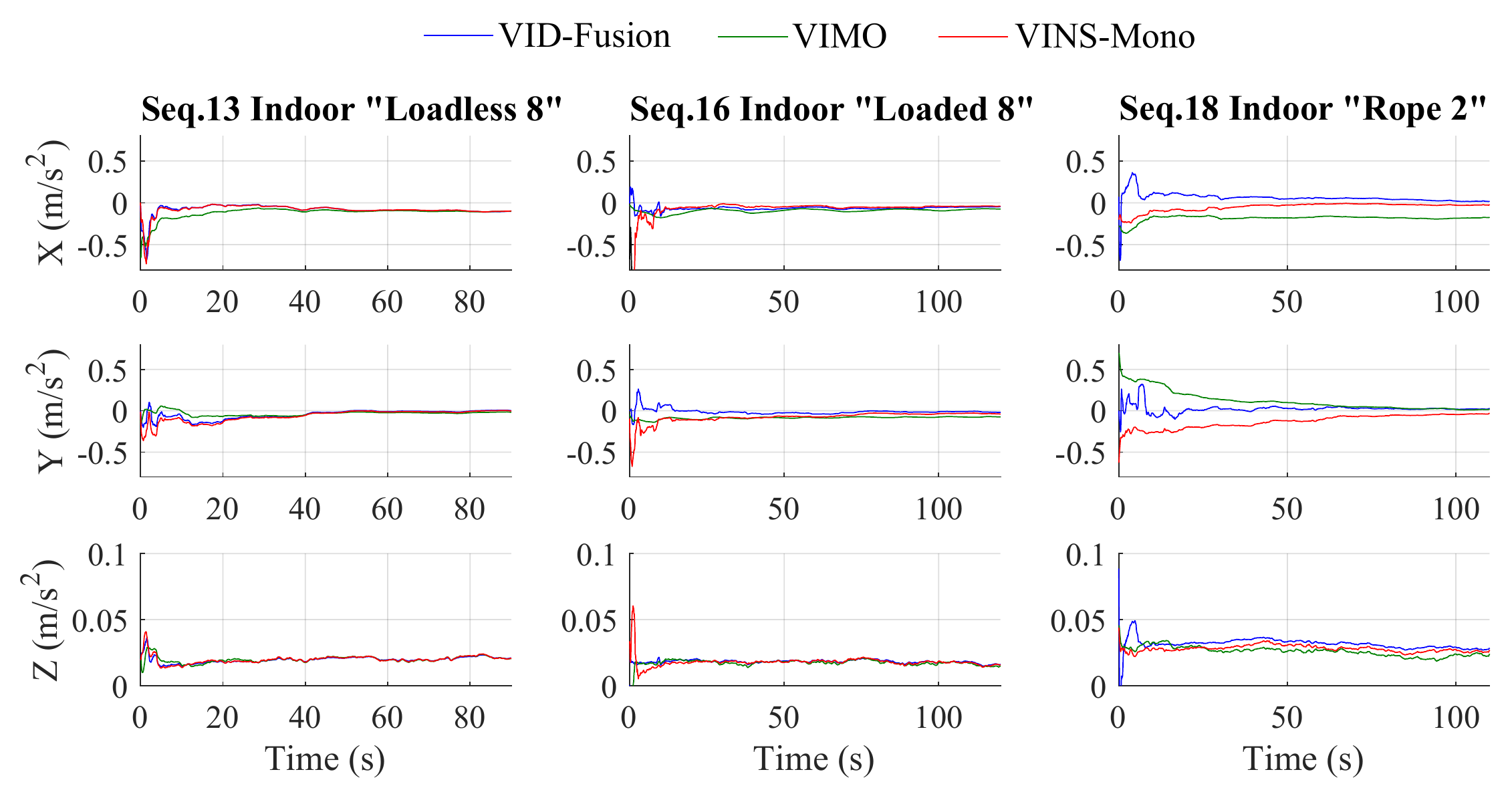}
		\caption{IMU acceleration bias estimation.}
		\label{bias_estimation}
	\end{center}
	\vspace{-0.8cm}
\end{figure}

To explore the two algorithms in more detail, 
Vins-Mono is chosen to compare their pose and acceleration bias estimation in the same scenarios.

\subsubsection{without load} 
The three algorithms have similar performance in pose estimation (Fig.~\ref{pose_estimation_3}-a) 
and all the estimated acceleration biases converge to the same value (Fig.~\ref{bias_estimation}-a).

\subsubsection{with a fixed load}
Although VIMO and VID-Fusion perform better in pose estimation (Fig.~\ref{pose_estimation_3}-b),
the former could not estimate external force correctly (Fig.~\ref{external_force_estimation_load}) 
with a different convergence of bias estimation (Fig.~\ref{bias_estimation}-b).

\subsubsection{external force changes} In the scenario 
where a rope is attached between the multirotor and the force sensor, 
there is little difference among the three algorithms 
in pose estimation (Fig.~\ref{pose_estimation_3}-c).
VIMO and VID-Fusion can't track the external force ideally (Fig.~\ref{external_force_estimation_sensor}), 
which may lead to biased estimation of acceleration bias (Fig.~\ref{bias_estimation}-c).

%

\section{Discussion and future work}
\label{sec:conclusion}
\subsection{Conclusion}
In this paper, we develop a multirotor platform carrying various time-synchronized sensors. 
Besides, we present the VID multirotor dataset for state and external force estimation. 
By several experiments, we validate the utilities of our proposed dataset 
and identify some problems with existing algorithms in external force estimation. 

\begin{enumerate}
\item
VIMO and VID-Fusion estimate a joint external force through the known dynamical model.
Once the mass and the thrust coefficients are provided inaccurately, they would be counted in the external force. 
\item
In both algorithms, the multirotor dynamics do not contribute significantly to pose estimation either in the absence of external forces or in the presence of a constant load.
And if the external force varies, it affects the bias estimation, 
or worse, would cause the estimator to crash when the external force changes too much, 
for example, during take-off or landing.
\item
It is not reasonable to treat all forces other than thrust as a combined external force, 
since there is such as drag force that can be modeled during flight.
\end{enumerate}

\label{sec:limitations}
\subsection{Limitations}
Our dataset still has several known limitations which need to be addressed in the near future.
\begin{enumerate}
\item It is unaggressive flight data, because of insufficient motor power.
\item It is inconvenient to provide the ground truth of external torques during flight.
\item It is abscent to provide more detailed and accurate model and its parameters instead of the individually identified propulsion units during flight.
\end{enumerate}

\label{sec:Future work}
\subsection{Future work}
It is our sincere hope that the VID dataset will facilitate scientific research and industrial applications in related fields.

In the future, we plan to improve the capabilities of drones 
by equipping more perceptual sensors (such as event camera or lidar), 
offering more detailed parameters, 
recording more aggressive and challenging sequences, 
providing ground truth of external torques which are lacking currently.

    \newlength{\bibitemsep}\setlength{\bibitemsep}{0\baselineskip}
    \newlength{\bibparskip}\setlength{\bibparskip}{0pt}
    \let\oldthebibliography\thebibliography
    \renewcommand\thebibliography[1]{%
        \oldthebibliography{#1}%
        \setlength{\parskip}{\bibitemsep}%
        \setlength{\itemsep}{\bibparskip}%
}

\bibliography{ICRA2022_KyZhang.bib}

\end{document}